\documentclass[lettersize,journal]{IEEEtran}
\usepackage{amsfonts}
\usepackage{amsmath}
\usepackage{algorithm} 
\usepackage{graphicx}
\usepackage{caption}
\usepackage{algpseudocode} 
\usepackage{multirow}
\usepackage{xcolor}
\newcommand\Tstrut{\rule{0pt}{2.6ex}}         
\newcommand\Bstrut{\rule[-0.9ex]{0pt}{0pt}}   

\begin{document}

\title{Robot Active Neural Sensing and Planning in Unknown Cluttered Environments}

\author{Hanwen Ren and Ahmed H. Qureshi
\thanks{The authors are with the department of computer science, Purdue University, West Lafayette, IN 47907 USA (email: ren221@purdue.edu; ahqureshi@purdue.edu)}}



\maketitle

\begin{abstract}
Active sensing and planning in unknown, cluttered environments is an open challenge for robots intending to provide home service, search and rescue, narrow-passage inspection, and medical assistance. Although many active sensing methods exist, they often consider open spaces, assume known settings, or mostly do not generalize to real-world scenarios. We present the active neural sensing approach that generates the kinematically feasible viewpoint sequences for the robot manipulator with an in-hand camera to gather the minimum number of observations needed to reconstruct the underlying environment. Our framework actively collects the visual RGBD observations, aggregates them into scene representation, and performs object shape inference to avoid unnecessary robot interactions with the environment. We train our approach on synthetic data with domain randomization and demonstrate its successful execution via sim-to-real transfer in reconstructing narrow, covered,  real-world cabinet environments cluttered with unknown objects. The natural cabinet scenarios impose significant challenges for robot motion and scene reconstruction due to surrounding obstacles and low ambient lighting conditions. However, despite unfavorable settings, our method exhibits high performance compared to its baselines in terms of various environment reconstruction metrics, including planning speed, the number of viewpoints, and overall scene coverage. 
\end{abstract}

\begin{IEEEkeywords}
Active sensing, scene reconstruction, planning and control, deep learning, unknown environments.
\end{IEEEkeywords}

\section{Introduction}
\IEEEPARstart{A}{ctive} sensing is a complex control problem where a robot with onboard sensors maximizes the information gained through interaction with an underlying environment \cite{ryan2010particle}. An efficient gathering of dense visual perception in the unfamiliar, narrow passage environments is a crucial subtask of active sensing for robots intending to assist people in their daily lives \cite{kidd2008robots}, providing services in search and rescue \cite{akin2013robocup}, or performing autonomous surgery \cite{shademan2016supervised}. For instance, consider an assistive robot at home retrieving a specific item. In this scenario, a robot must search various places, including cabinets with narrow passages and limited lighting. Similarly, in search and rescue, especially at disaster sites such as struck by earthquakes or underwater, the environment will be unknown \cite{cao2019target}, and service robots would have to actively sense the scene and perform rescue operations. A teleoperated assistive robot surgery also often requires endoscopic cameras to create a dense perception for the far distant surgeons to function efficiently \cite{enayati2016haptics}.\par

Recent advancements in deep learning have opened ways for model-free, scalable robot planning and control with tools like 3D convolution neural networks (3DCNN) \cite{ji20123d}, PointNet++ \cite{qi2017pointnet++}, and Transformers \cite{vaswani2017attention} that have been employed in various robot locomotion \cite{yang2021learning}, visuomotor control \cite{yuan2021dmotion}, and planning tasks \cite{johnson2021motion}. Another recent method NeRP \cite{DBLP:conf/rss/QureshiMPYF21} introduced Model Predictive Control (MPC) style algorithm for solving rearrangement tasks with unknown objects from raw point-cloud observations. Inspired by these developments, this paper presents an efficient, fast deep learning-based active sensing technique for robot manipulators with an in-hand camera to reconstruct the given narrow, cluttered environments with arbitrary unknown objects. 

Our method explores the given setting with minimal camera viewpoints while maximizing the overall information gain. It introduces novel deep neural network-based next best viewpoint generation algorithms that take the existing scene representation and output the next best viewpoint for a robot manipulator to interact and observe the underlying complex environment without collision. Our framework actively gathers the visual RGBD observations from the given viewpoints, registers them into scene representation, and infers the unknown object shapes from their partial observations to avoid unnecessary robot interaction with the given environment during scene reconstruction.
The main contribution and salient features of our active neural sensing approach are summarized as follows.
\par
\begin{figure*}[h]
    \centering
    \includegraphics[trim = {0cm 16cm 0 0cm}, clip, width = 16.5cm]{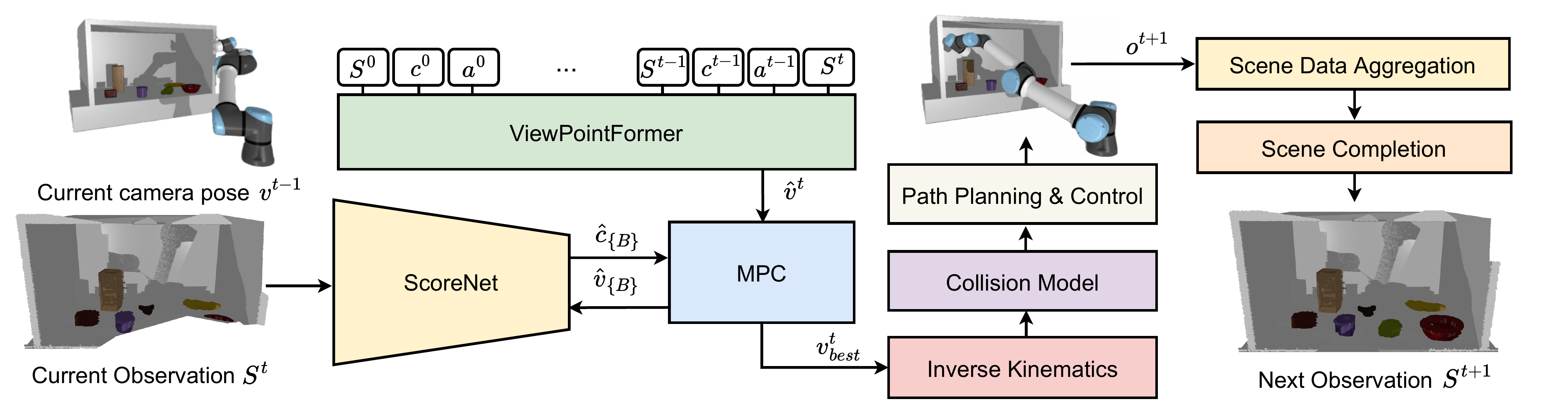}
    \caption{Active Neural Sensing System: Starting from the initial scene representation $S^t$, our active viewpoint generation module finds the next best viewpoint $\hat{v}^t_{best}$ using either VPFormer or bilevel MPC framework, guided by the ScoreNet function. The inverse kinematic model computes the robot configuration $q^t \in \mathcal{Q}$ corresponding to the best viewpoint. Next, the path planning and control module moves the robot to the desired viewpoint using our collision model approximation of the observed environment to avoid obstacles. After reaching the proposed viewpoint, the new information $o^{t+1}$ is segmented and aggregated to the existing scene representation. Finally, our method performs shape inference to complete partially observed objects, resulting in $S^{t+1}$. The new scene representation becomes the input to our framework at the next step, and the process continues until the desired coverage score is achieved.}
    \label{fig1:overview}
\end{figure*}
\begin{itemize}
   \item 3DCNN-based score function that takes the scene representation based on past observations and a viewpoint candidate to forecast possible scene coverage, thus guiding viewpoint planning and preventing unnecessary robot interaction with the given environment. 
   \item Bilevel MPC style next best viewpoint generation algorithm that leverages our score function and converges to viewpoint sequences leading to maximum scene coverage. 
  \item Transformer-based viewpoint sequence modeling and generation for fast scene reconstruction, resulting in a small number of viewpoints and almost real-time maximum scene coverage. We train this module using data from our bilevel MPC algorithm guided by our score function under an offline reinforcement learning paradigm \cite{levine2020offline}.
  \item Scene registration framework combining the information from various viewpoints into a unified scene representation and performing neural scene inference to complete object shapes from their partial 3D point clouds. The scene completion also prevents unnecessary interaction steps, thus reducing the number of viewpoints required for maximum scene coverage.
 \item A strategy to approximate collision models for the underlying planning and control modules to prevent robot's self-collisions and collisions with the unknown, partially observed environments during motion execution to reach the given viewpoints for active sensing.
  
 \item A unified fast active sensing framework combining viewpoint generation and robot control methods for scene reconstruction with results demonstrated in complex simulation and real-world cabinet-like environments using a 6-DOF manipulator with an in-hand RGBD camera.
 \end{itemize}

\section{Related Work}
\noindent The most relevant work to our approach can be categorized into classical search-based approaches, teleoperated robot sensing, and modern data-driven methods. The classical active sensing methods mainly use frontier-based or sampling-based methods to find the next best viewpoint. In the frontier-based approaches \cite{yamauchi1997frontier, vasquez2014volumetric, border2018surface}, the viewpoints are generated on the detected surface while ensuring certain amount of visual overlap with the previous observation. Therefore, these methods require many viewpoints for scene coverage. Furthermore, the viewpoint stitching strategy limits the applicability of such techniques to narrow passages and cluttered environments. 
In the sampling-based methods, the planners generate viewpoint sequences using a random search and move to the best viewpoint achieving the highest utility function value \cite{vasquez2014view, bircher2016receding}. In a similar vein, \cite{bircher2015structural} introduces the iterative viewpoint re-sampling to compute the best viewpoint sequences with low-cost connections by conducting multiple phases of sampling and cost estimation. However, most of these methods assume the mesh representation of the environment is given in advance at test time, i.e., with known object categories and underlying geometries. In addition, their reliance on the ray tracing technique also limits their generalization to unknown, cluttered scenarios since it is under the assumption that all space inside the proposed viewpoint's field of view is fully observable.\par
The teleoperation-based active sensing chooses the next step based on the selection of the human operator \cite{delgorge2005tele, xiao2021autonomous, abolmaesumi2002image}. These methods are primarily used in distant, risky, or clinical environments where direct human interaction is infeasible. By projecting human presence in the remote environment, it provides an effective solution to satisfy field demands while reducing any potential challenges due to direct human engagement. However, one of the major drawbacks of the teleoperated method is that human operators cannot get adequate situational awareness because of robot sensing limitations, such as a limited field of view \cite{xiao2021autonomous}. As a result, the selected viewpoint may not be close to optimal. A possible way to solve such issues is by introducing multiple robots and their operators to gather information \cite{kiribayashi2018design}, which generally is not cost-effective and requires an extra workforce. On the other hand, no automation is involved in these active sensing methods, making them less likely to be deployed on larger scales.\par
Latest applications incorporate data-driven methods into the active sensing paradigm. Neural networks are introduced to replace traditional utility functions and directly predict information gain without physically moving to generated viewpoints \cite{zeng2020pc, mendoza2020supervised, sun2021learning}. However, these methods are primarily evaluated in simulations with virtual cameras and no sim-to-real transfer to deal with real-world sensory and actuation noise. Furthermore, only single object reconstruction is performed, making them inapplicable to real-world, cluttered, narrow passage environments that usually have multiple objects with limited visibility. There also exist data-driven approaches that perform real-world experiments \cite{lepora2019pixels, sucar2020nodeslam, peters2022robot, engin2020higher}. However, they usually consider a single object reconstruction task and place the robot in open spaces without significant obstacles hindering robot motion during active sensing. 
\section{Active Neural Sensing}
\noindent This section formally presents the main components of our robot active neural sensing and planning, namely Active viewpoint generation, collision modeling for path planning, and Scene Registration.

\subsection{Problem Definition}
\noindent Let a given environment be denoted as $S \in \mathbb{R}^3$, with $S_o$ and $S_{ou}=S\backslash S_o$ being its observed and unobserved regions, respectively. The ranges of $S$ along $x,y,$ and $z$ axis are assumed to be known and denoted as $d_x, d_y,$ and $d_z$, respectively. The value of a coordinate $S(i,j,k)$ is either 1 or 0, where the former indicates the observed and the latter indicates the unobserved points. Let $\phi$ be a coverage measure that determines the volumetric ratio of the observed region $S_o$ over the entire scene $S$ as: 
\begin{equation}
    \phi(S_o, S) = \frac{\sum_i^{d_x}\sum_j^{d_y}\sum_k^{d_z} S(i,j,k) \neq 0}{d_x\cdot d_y\cdot d_z}
\end{equation}
Furthermore, we assume a setting where the robot arm is equipped with an in-hand camera. Let an in-hand camera pose be denoted as $v \in \mathcal{V}$ and its corresponding robot arm configuration as $q\in \mathcal{Q}$, where $\mathcal{V} \in \mathbb{R}^7$ is a 7D camera pose space and $\mathcal{Q} \in \mathbb{R}^d$ is a $d$-dimensional robot joint space. An inverse kinematics function gives the mapping from in-hand camera pose space $\mathcal{V}$ to robot configuration space $\mathcal{Q}$. At a given viewpoint $v$, the system makes an RGBD visual observation $o \in \mathcal{O}$, where $\mathcal{O}$ is a visual space.  Our observed scene $S^t_o$ at step $t$ is a function of in-hand camera views integrated from time $0$ to $t$, i.e., $\{o^0, o^1,...,o^{t}\}$. 
Let a policy $\pi: S \times \mathcal{V}  \mapsto \mathcal{O}$ be a function that takes the current scene representation $S^t$, at time $t$, and outputs the next in-hand camera viewpoint $v^{t}$ leading to the next scene $S^{t+1}$. Since $S^{t+1}$ is constituted from the policy action $v^t$ and its resulting observation $o^{t+1} \in \mathcal{O}$, we denote the next scene as $S^{t+1}(v^t)$ to describe our objective function. 
Our objective is to find a policy function $\pi$ that outputs the reachable, shortest path sequence of in-hand camera viewpoint poses $v \in \mathcal{V}$ to maximize the overall scene coverage $\phi$, thus forming a min-max formulation as:
\begin{equation}
    \max_{v^t \sim \pi} \min_T \sum_{t=0}^{T-1} \phi(S_o^{t+1}(v^t)\backslash S_o^t, S)
\end{equation}

The reachability constraint implies that the generated viewpoints and their corresponding robot configurations must be kinematically feasible and collision-free, allowing the underlying robot motion planner and controller to create feasible motion sequences for obtaining visual observations. For the remainder of this paper, we introduce few more notations for brevity. Let any arbitrary set or list $a$ with $B$ number of elements be denoted as $a_{\{B\}}$. The scene representations will be denoted as $S^{t+1}, S^{t+1}_o,$ and $S^{t+1}_{ou}$ resulting from the action $v^t \in \mathcal{V}$ taken by the policy $\pi$ at time step $t$.  

\subsection{Active Viewpoint Generation}
\noindent \subsubsection{ScoreNet}
\noindent Viewpoint generation algorithms for active sensing must follow foresight to find the best strategy for exploring the unknown environment while avoiding unnecessary interaction steps. Thus, we introduce ScoreNet, a neural network-based approach to forecast the overall scene coverage at various viewpoints over the observed space without physically moving the in-hand camera manipulator. Our ScoreNet function $f_\theta$ with parameters $\theta$ provides a real-valued scene coverage foresight $\hat{c} \in \mathbb{R}$ based on the current observation $S^t$ and the next viewpoint candidates $v^t_{\{N\}}$ proposed by the generator $\pi$, i.e.,
\begin{equation}
   \hat{c}_{\{N\}} \leftarrow f_{\theta}(S^{t},v^t_{\{N\}}) 
\end{equation}
During execution, our scoring function evaluates the proposed viewpoint candidates and assists the underlying planning algorithm in selecting the best viewpoint sequences leading to maximum coverage in the least number of interaction steps. We train the Scorenet in a supervised manner using the Mean Squared Error (MSE), i.e., $\frac{1}{N} \sum_{i=1}^N \lVert c - \hat{c} \rVert^2$, between the predicted coverage $\hat{c}$ and the ground truth labels $c$ obtained using the Equation 1 on the randomly generated synthetic data. Our ScoreNet architecture comprises 3DCNN \cite{ji20123d} kernels to embed the scene representation $S$ into a latent space $Z_S$ and a fully-connected neural network for encoding the viewpoint $v$ into its embedding $Z_v$. These latent features are concatenated and transformed into the desired output $\hat{c}$ using another fully-connected neural network function.
\subsubsection{Active Viewpoint Generation Algorithms}
After the ScoreNet is trained, one of the remaining challenges is to generate and select the next best viewpoint among all the possible candidates. Thus, we design the following two active viewpoint generation methods.
\paragraph{Bilevel MPC Viewpoint Generation}
We propose an MPC-style viewpoint generation algorithm. A traditional MPC-style method \cite{qin2003survey} iteratively optimizes the sampling distribution, usually Gaussian $\mathcal{N}(\mu, \sigma)$, with parameters mean $\mu$ and standard deviation $\sigma$. It generates a variety of samples from $\mathcal{N}(\mu, \sigma)$ and selects the top $K_m$ elite samples through sorting based on some score function. Then, these elite samples are used to compute new distribution parameters, and the process of sampling and sorting continues for a certain number of iterations until the distribution parameters converge to some local optima. Although such approaches have been demonstrated to solve many problems, they are not ideal for viewpoint generation. 

The viewpoint generation using traditional MPC is an ill-posed formulation since there can be multiple viewpoints scattered around the given environment that can lead to similar coverage scores from the ScoreNet. This prevents standard MPC from converging and often results in high variance approximates. To overcome these issues, we propose a bilevel MPC algorithm for viewpoint generation. In the first stage, we generate a variety of random viewpoint samples from observed space $S^t_o$ and select the best viewpoint with the highest score given by our ScoreNet, i.e., 
\begin{equation}\label{method1}
    \mu =  \arg \max_{v_{s} \in S_o^t} f_{\theta}(S^{t}, v_{s} )
\end{equation}
In the second stage, we define a distribution around the best viewpoint $\mu$ and apply the traditional MPC style optimization approach to arrive at distribution parameters leading to elite samples for maximum coverage. Note that the second stage refines the viewpoint from the first stage since the random sampling technique may not converge to the optimal local region.

Algorithm 1 outlines our bilevel MPC method. It begins by finding the best viewpoint $\mu$ through random sampling points $v_{\{N_{mpc}\}}$ from a scene $S^t_o$ and passing them through the pre-trained ScoreNet to compute their corresponding predicted coverage scores $\hat{c}^t_{\{N_{mpc}\}}$. Once the best viewpoint guesses $\mu$ is selected with the maximum coverage score, the local samples are generated using pre-defined $\sigma$. The initial guess $\mu$ and variance $\sigma$ are then further refined through our second phase MPC, as shown in Line 3-7. The $\mathrm{Feasible}$ function ensure all sampled viewpoints satisfy the kinematic feasibility and collision-avoidance constraints. The latter condition is satisfied with our collision model approximated from the point cloud of observed space as described later in this section. The $\mathrm{Sort}$ function rearranges the list of sampled viewpoints in descending order concerning score values such that the highest scoring viewpoint appears first. Finally, the $\mathrm{Fit}$ updates the mean $\mu$ and variance $\sigma$ from the elite samples to update the Gaussian distribution for sampling and sorting in the next iteration.
\begin{algorithm}
	\caption{MPC Based Viewpoint Generation Algorithm} 
	\label{alg: MPC algo}
	\begin{algorithmic}[1]
	    \State initialize $\sigma$
	    \State $\mu \leftarrow \arg \max_{v_s \in S^t_o} f_{\theta}(S^{t}, v_{s} )$ \Comment{initialize mean}
	    \For {$iter \leftarrow 1$ to $N_{iter}$}
	            \State $v^t_{\{N_{mpc}\}} \sim \mathrm{Feasible}(\mathcal{N}(\mu, \sigma), S^t_o)$ \Comment{ batch sampling}
	        \State $\hat{c}^t_{\{N_{mpc}\}} \leftarrow f_{\theta}(S^{t}, v^t_{\{N_{mpc}\}})$ \Comment{viewpoint scores}
	        \State $(v^t, \hat{c}^t)_{\{K_m\}} \leftarrow \mathrm{Sort}(v^t_i, \hat{c}_t^i)[:K_m]$ \Comment{elite samples}
	        \State $\mu, \sigma \leftarrow
	        \mathrm{Fit}(v^t_{\{K_m\}})$ \Comment{update parameters}
	    \EndFor\\
	    \Return $(v^t, \hat{c}^t)_{\{K_m\}}$ \Comment final result
	\end{algorithmic} 
\end{algorithm}
\paragraph{Transformer-based Viewpoint Generation} 
In this module, we introduce a neural-informed viewpoint generator that quickly finds a sequence of viewpoints leading to maximum scene coverage without requiring an exhaustive, time-consuming viewpoint random search during execution. We learn our neural function through an offline reinforcement learning paradigm \cite{levine2020offline} guided by the trajectory sequences obtained from our bilevel MPC-based viewpoint generator. To best capture the causal relationship within sequential data of our scene observation $S^t$, viewpoint action $v^t$, and coverage reward $c^t$ pairs, we leverage the multi-head self-attention mechanism of the Transformer models \cite{vaswani2017attention}. Therefore, we name our neural method as ViewPointFormer (VPFormer). 

In our setup, each sequence token is composed of three elements $(c^i, S^i, v^i)$, indicating coverage rate, scene, and viewpoint at step $i \in \mathbb{N}$. The scene observation $S^i$ is encoded to $Z^i_{S}$ by 3D convolution kernels while $c^i$ and $v^i$ are encoded into $Z^i_{c}$ and $Z^i_{v}$ through fully connected neural network layers. The concatenation of $Z^i_{c}$, $Z^i_{S}$ and $Z^i_{v}$ forms latent token representation $Z^i$. A sequence of latent token representations are computed until the current time step $t$, i.e., $Z=\{Z^0,\cdots,Z^t\}$. In addition, the positional encodings \cite{vaswani2017attention} are applied to $Z$, so that the sequence order is preserved in the latent token representations. These token embeddings are passed through several Transformer encoder blocks. Inside each encoder block, multi-head self-attention is performed on $Z$. For each head, the Transformer parameters known as queries $Q$, keys $K$, and values $V$ are obtained through linear projections of $Z \in \mathbb{R}^{d_z}$ using learnable weight matrices $W^Q $, $W^K$, and $W^V$, respectively. The dimension of each weight matrix is $\mathbb{R}^{d_Z \times d_k}$, where $d_z$ is the dimension of latent encoding $Z$ and $d_k$ is the size of the projection embedding. The attention coefficients are calculated by the inner product of $(Q, K)$, which are then combined with the values $V$ to get the self-attention matrix of the given sequence, i.e., 
\begin{equation}
    \mathrm{Attention}(Q, K, V) = \mathrm{softmax}(\frac{QK^T + M}{\sqrt{d_k}})V,
\end{equation}
A binary mask $M$ is also imposed to prevent the tokens arriving later in the sequence from accessing the earlier embeddings. Since we use multi-head attention, the self-attention parameters from all heads are concatenated and transformed into viewpoint action $\hat{v}^t$ using a fully-connected neural network. {\color{red} }Our multi-head self-attention captures different latent dependencies in the given sequence and therefore generates action $\hat{v}^t$ that takes all past scene representations, actions, and coverage rates into consideration for viewpoint planning.
In our offline RL setup, we define the training loss as MSE between the predicted viewpoints $\hat{v}$ and the ground truth viewpoints $v$ provided by our bilevel MPC.
Furthermore, at any step during execution, a new viewpoint is generated autoregressively based on all previous information comprising coverage rates, scene representations, and viewpoints. To further refine the predicted action, we sample several viewpoints around $\hat{v}^t$ with fixed variance $\sigma$ and output the best viewpoint $v^t$ that gets the highest predicted coverage score from ScoreNet.\par
\subsection{Collision Modelling for Path Planning}
\noindent Our path planner moves the robot from its current position $v^{t-1}$ to the next desired viewpoint $v^t$ while satisfying the kinematic and collision-avoidance constraints. In our setup, we assume an in-hand camera of a robot manipulator for active sensing. Therefore, we convert the in-hand camera pose $v^t$ to the robot end-effector pose through an offset transformation matrix and determine the corresponding robot joint configuration $q^t \in \mathcal{Q}$ using an inverse kinematics approximator \cite{d2001learning}. Furthermore, to satisfy the collision constraints while moving from the previous configuration $q^{t-1}$ to $q^t$, the robot arm must prevent collision with itself, observed objects, unobserved space, and the boundary of the scene at all times. These collisions can be avoided by adequately defining a surrounding collision model. 

To build a robot collision model, we define a function that maps robot joint angles in C-space, i.e., $q^t$, to the poses of each link $(p_{l^t})_{\{d\}}$. Then self-collision can be detected by checking contacts between non-adjacent links after placing their collision meshes at the corresponding poses $(p_{l^t})_{\{d\}}$. The collision models for detected objects are obtained by calculating the smallest bounding box of object point clouds. We build the scene boundary collision model by placing thin rectangle boxes at all boundary surface locations. Note that, at any step $t$, we also want to prevent the robot from occupying any unknown space $S_{ou}^{t}$. Thus, a surface mesh of $S_{ou}^{t}$ is built using a ball-pivoting algorithm \cite{bernardini1999ball}, which serves as its collision model. Finally, we incorporate our collision models into the Flexible Collision Library \cite{pan2012fcl} for an underlying motion planner to check the robot's self-collision and environment collision during motion generation. We use RRT* \cite{karaman2011sampling} as our motion planner from the Open Motion Planning Library \cite{moll2015benchmarking-motion-planning-algorithms} .
\subsection{Scene Registration}
\noindent In this section, we describe our scene registration process that takes the new observation $o^{t+1}$ and scene representation $S^t$ and passes them through multiple steps ranging from instance segmentation and information aggregation to object shape completion, resulting in updated scene representation $S^{t+1}$.

The camera at a viewpoint $v^t$ outputs an RGBD image as the new observation $o^{t+1}$. First, we process $o^{t+1}$ by applying the instance segmentation algorithm Dectectron2 \cite{wu2019detectron2} to extract each object's RGBD information. This results in object-centric partial point clouds denoted as $\boldsymbol{x}_{l}$ with $l$ indicating the number of instance segmentations. Next, we transform the extracted point clouds from the image frame to the global scene frame using the robot in-hand camera intrinsic and extrinsic parameters.  

After the instance segmentation and projection to the global scene frame, we combine the new observed information $\boldsymbol{x}_{l}$ with the previous scene representation $S^t$ to generate the next scene $\hat{S}^{t+1}$. The camera observations are likely to overlap, meaning they may capture the same object multiple times from different angles. Therefore, to avoid repetition, our method registers the newly observed object point clouds $\boldsymbol{x}_{l}$ to their corresponding existing object point clouds in the scene representation $S^t$. Our approach assumes that if two sets of point clouds share the same point in the scene space, they belong to the same object instance. Therefore, for each new object point cloud observation, we check its minimal distance from all existing object point clouds. If the minimum distance between any new and old object point clouds is below a predefined threshold $\eta \in \mathbb{R}$, they are appended together as the same instance; otherwise, a new object instance is created in the scene representation. Furthermore, we also identify and aggregate empty spaces in the old scene representation $S^t$ and new observation $o^{t+1}$. We use a pinhole camera model \cite{kannala2006generic} along with the camera intrinsic parameters to find scene areas in the camera's visibility cone. By further comparing their distance to the camera lens and the corresponding depth measures, we label those areas as either observed and empty or unobserved.

Once new information is appended to scene $S^t$ leading to new scene $\hat{S}^{t+1}$, we perform the object-wise scene completion. Since our in-hand camera has a limited field of view, the objects are usually partially observed and require a large number of viewpoints to reconstruct the given scene. In contrast, the objective of our proposed method is to construct the scene using a minimal number of viewpoints to avoid unnecessary interaction with the environment for semantic scene understanding tasks. Therefore, we apply the object-wise shape completion algorithm to $\hat{S}^{t+1}$ for maximizing the scene coverage using only the available information.

Our scene completion module uses a Transformer-based geometry-aware shape completion network to infer the missing parts of detected objects in the scene. Our method is based on the PoinTr framework \cite{yu2021pointr}. However, we introduce several modifications to PoinTr so that our approach can tackle scenes with unknown objects. In the original PoinTr design, the objects' sizes are assumed to be known as prior information to project the partial point clouds into a unit sphere. Furthermore, the PoinTr computes the unit sphere by subtracting the object center position from the partial point cloud and dividing it by the known object volume. These point clouds in the unit sphere are then passed through a neural network for shape inference. However, in practice, under the unknown environment setting, especially during active sensing, the objects' sizes cannot be known a priori. Therefore, we set the normalization constant to the maximum possible volume of the household objects. Moreover, we compute the mean of the partial point cloud as a subtraction term for projection to the unit sphere. During training, the objective of the shape completion network is to minimize the L2 Chamfer Distance between the ground truth point cloud $g$ and the inferred point cloud $\hat{p}$:\\
\begin{equation}\label{comp_loss_function}
    L = \frac{1}{|\hat{p}|}\sum_{i\in \hat{p}}\min_{j\in g}||i-j||^2 + \frac{1}{|g|}\sum_{j\in g}\min_{i\in \hat{p}}||j-i||^2
\end{equation}
Once trained, our shape completion network further processes the object point clouds in scene $\hat{S}^{t+1}$ to infer their shapes thus leading to the next observation $S^{t+1}$ for our underlying viewpoint generation and selection algorithms.

\begin{algorithm}[h]
	\caption{Active Neural Sensing Approach} 
	\label{alg: system overview}
	\begin{algorithmic}[1]
	    \State $S^0_o \leftarrow \emptyset$; $c \leftarrow 0$; $t \leftarrow  0$; $\alpha^0 \leftarrow \emptyset$;
	    $\sigma \leftarrow \emptyset$
	    \Comment{Initialization}
        \State $f_\theta$ \Comment{Initialize ScoreNet function}
        \State $\pi_\Phi \leftarrow $ $\{$Bilevel MPC, VPFormer$\}$
     \While {$c \leq c_{max}$ and $t \leq T_{max}-1$}
	        \State $\alpha^t \leftarrow \mathrm{CollisionModel}(S^t)$ 
	        \State $v^t_{\{B\}} \leftarrow \pi_\Phi(S^{t}_o)$ \Comment{next viewpoints}
	        \State $\hat{c}^t_{\{B\}} \leftarrow f_\theta(S^{t},v^t_{\{B\}})$ \Comment{next view scores}
		  \State $(v^t, \hat{c}^t)_{\{B\}} \leftarrow \mathrm {Sort}(v^t_i,\hat{c}^t_i)$ \Comment{sort viewpoints}
		  \State $j \leftarrow 0$ \Comment{set iterator}
		  \State $v_\mathrm{best} \leftarrow \emptyset$ \Comment{best view}
	        \While {$\mathrm{notValid}(\sigma)\; \boldsymbol{\mathrm{and}}\;  j<B$} 
	             \State $q_t \leftarrow \mathrm{JointAngle}(v^t_j)$
	             \State $\tau \leftarrow \mathrm{ComputePath}(q^{t-1}, q^t,  \alpha^t)$
	             \State $v_{best} \leftarrow v^t_j$
	             \State $j \leftarrow j+1$
	        \EndWhile
	        \State Move robot using $\tau$ and obtain $v_{real} \: \mathrm{and} \: o^{t+1}$
	        \If {$v_{real} \neq v_{best}$} \Comment{validate result viewpoint}
	            \State continue
	        \Else
	            \State $S^{t+1}_o \leftarrow \mathrm{SceneRegistration}(o^{t+1})$ 
	            \State $c \leftarrow \phi(S^{t+1}_o, S)$ \Comment{update coverage rate}
	       	   \State $t \leftarrow t + 1$ \Comment{prepare for next step}
	        \EndIf
	    \EndWhile\\
        \Return $S^{t+1}_o$ \Comment{final result}
	\end{algorithmic} 
\end{algorithm}
\subsection{Active Neural Sensing Algorithm}
\noindent Algorithm \ref{alg: system overview} describes our active neural sensing via robot manipulation system to efficiently explore unknown environments. Our system starts at step $t = 0$ with scene representation $S^0$, where the observed space $S^0_o$ is an empty set while the unobserved space $S^0_{ou}$ fills up the entire scene. Thus, the initial scene coverage score $c$ is set to 0. A robot arm with an in-hand camera is placed in front of the scene at a random location. Our trained ScoreNet predicts the coverage score $\hat{c}$ given a viewpoint candidate $v$ and a scene representation $S$ to guide the viewpoint generation policy $\pi$.\par
In each time step $t$, the collision model $\alpha^t$ is built based on the most up-to-date information about the scene $S^t$ to avoid the robot's self-collisions and collisions with the surrounding environment. Depending on policy preference, a set of collision-free viewpoint candidates $v^t_{\{B\}}$ can be generated from the bilevel MPC or VPFormer algorithm. The bilevel MPC algorithm generates viewpoints iteratively, optimizing a sampling distribution. While VPFormer predicts the best viewpoint $\hat{v}^t$ directly from past coverage scores $c_{\{t-1\}}$, observations $S_{\{t\}}$, and viewpoints $v_{\{t-1\}}$. A set of viewpoints is sampled round $\hat{v}^t$ to form $v^t_{\{B\}}$ for which the ScoreNet predicts the coverage scores $\hat{c}^t_{\{B\}}$. The viewpoint candidates $v^t_{\{B\}}$ and their coverage scores $\hat{c}^t_{\{B\}}$ are sorted pairwise in the descending order of coverage values to prioritize the selection of viewpoints with the best score in the execution process (Line 5-8).\par
Our sorted list comprises viewpoints and their scores pairs $\{(v^t_0, \hat{c}^t_0), (v^t_1, \hat{c}^t_1),...,(v^t_B, \hat{c}^t_B)\}$. Using the inverse kinematics algorithm, we start from the viewpoint $v^t_0$ and calculate its corresponding robot arm pose $q^t_0$. If $q^t_0$ exists, our path planner computes the robot motion trajectory from $q^{t-1}$ to $q^t_0$ under collision avoidance constraints governed by $\alpha^t$. If there does not exist a feasible path solution to reach $q^t_0$, we move to the next viewpoint in our list, and the path finding process is repeated until a valid viewpoint $v_{best}$ is found for which there exists a valid path solution $\tau$ (Line 9-15).\par
Next, we execute the robot motion trajectory $\tau$, and the robot arrives at a viewpoint denoted as $v_{real}$. We compute the Euclidean distance between $v_{real}$ and $v_{best}$ before observing through the camera. If the error is above a threshold, the viewpoint is ignored otherwise, a new observation $o^{t+1}$ is generated for the scene registration process. The scene registration performs the instance segmentation, data aggregation, and objects' shape completion based on $o^{t+1}$. The newly obtained information is added to $S_o^t$ leading to next scene observation $S_o^{t+1}$. Therefore as the algorithm proceeds, the size of $S_{ou}^{t+1}$ shrinks since it is a complement of $S_o^{t+1}$. Before entering the next time step $t+1$, the actual scene coverage score is computed using our function $\phi$ (Line 18-23). \par
Lastly, our method continues iteratively to generate the following best viewpoints and perform corresponding robot motions for making new observations. The process repeats until the actual scene coverage exceeds our minimum desired coverage threshold $c_{max}$ or if it reaches the iteration limit $T_{max}$, resulting in the final reconstructed scene $S_o^{t+1}$ (Line 4 \& 26). 
\begin{figure*}[ht]
    \centering
    \includegraphics[trim = {1cm 0cm 0cm 0cm}, clip, width = 18cm]{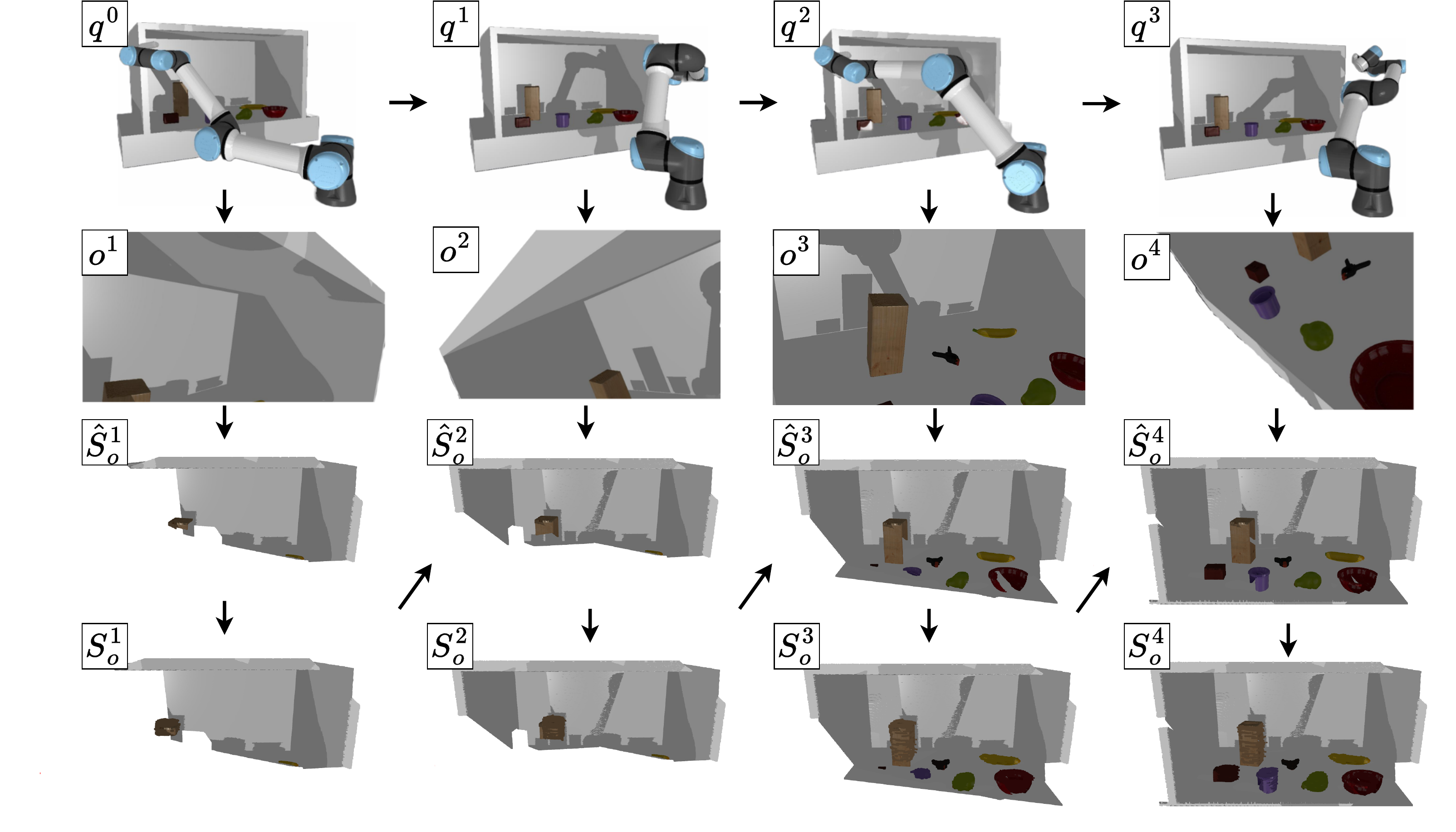}
    \caption{A viewpoint sequence generated by our VPFormer-based  active sensing framework in a narrow, cluttered environment. Rows 1-4 exhibit the robot viewpoint configurations, in-hand camera observations, scene representations, and scene after object shape completion. Each column's results come from the same viewpoint at a specific time step. The final reconstructed scene is shown in bottom right $S^4_o$.}
    \label{fig:sim_figure}
\end{figure*}

\section{Implementation Details}
\noindent This section describes our simulation and real-world environment settings, model architectures, hyperparameters, and other training and testing dataset details. Furthermore, we present all measurements in meters unless stated otherwise.
\subsection{Environment Setup:}
\noindent In both simulation and real environment setups, we use the 6-DOF Universal Robots UR5e manipulator combined with the Intel Realsense D435i Depth camera as our active sensing platform. The camera is connected to the robot's end-effector using a 3D-printed mount bracket. The camera and the robot's end-effector share the same rotation but have a relative offset of $(0.11, 0, 0.07)$. The camera's horizontal field of view is 70.25 degrees, and its RGBD image solution is $1280\times720$. The models of household objects are from the YCB dataset \cite{calli2015benchmarking}, comprising objects ranging from utensils to fruit and grocery items of different shapes and sizes.
\subsubsection{Simulation setup} 
We set up our virtual environments in Nvidia Isaac Gym Simulator \cite{liang2018gpu}. Each generated scene is of an enclosed rectangle shape with one opening side, representing a cabinet-like scenario. We randomly place various household objects inside the cabinet scenario and situate the robot arm in front of it for active sensing. Furthermore, we randomize the following features of our environment setup to add domain randomization, allowing sim-to-real transfer of our trained neural functions. The scene dimensions $dx$, $dy$, and $dz$ are randomly selected from ranges $[0.5, 1]$, $[1.1, 1.5]$, and $[0.3, 0.7]$, respectively. Therefore, the scene center locates at $(0.3 + dx/2, 0)$ on the $x$-$y$ plane.  Furthermore, the cabinet height from the ground is also randomized within the range $[0.05, 0.2]$. The robot arm's base's geometric center is selected randomly along the x and y plane from $[-0.1, -0.5], [-0.3, 0.3]$. Finally, 3 to 10 objects from the YCB dataset are selected and placed randomly inside the cabinet, thus resulting in unique simulated scene setups. The objects' placement locations and orientations are randomized by dropping objects from above to the cabinet's surface.
\subsubsection{Realworld setup} 
To show the performance of our system in real-world scenarios at test time, we use a standard household cabinet with dimensions $(dx, dy, dz) = (0.65, 1.12, 0.66)$. The robot arm is placed on a stand with 0.95  offset from the ground, whereas the cabinet bottom surface height is 1.10 from the ground. Furthermore, we use real-world YCB dataset objects, including mug, bowl, and dummy fruits, to populate our environment. We also randomize the objects, their placements, and the relative robot's location from the cabinet to form the test dataset of ten environments. Note that we train our models in the simulation with domain randomization and use our real-world setup only for evaluation purposes to demonstrate the sim-to-real transfer of our framework.
\subsection{Data Generation: Training \& Testing}
\noindent A total of $15,203$ sample pairs of the form $(S, v, c)$ were generated in simulated environments to train and evaluate ScoreNet. We created 1000 scenes and sampled various random viewpoint sequences in each setting, resulting in trajectories of the form $\{S^0,v^0,c^0, \cdots, S^T,v^T,c^T\}$. The trajectories were broken into pairs $(S, v, c)$ and were randomly shuffled to break sequence correlation and prevent model overfitting during training. In the case of VPformer, we gather optimal viewpoint sequences using our bilevel MPC in simulated environments to train and evaluate our VPFormer. We store the scene representation $S$, viewpoint $v$, and coverage rate $c$ at each time step, resulting in one hundred seventy-five expert sequences. Finally, the shape completion network data was generated using the mesh files from the YCB dataset. We extracted mesh vertices and sampled 8000 points for each object using the farthest point sampling (FPS) method \cite{moenning2003fast}. To mimic realistic camera view settings, we performed random cuts of each object point cloud and formed partial and complete object point cloud pairs, resulting in a dataset of 1872 samples. From all data samples, we use 80\% for training and the remainder for evaluation. In addition, we also evaluate our models in the never-seen-before real setups. 
\subsection{Network Architecture and hyperparameters}
\noindent This section describes our model architectures implemented in Pytorch \cite{NEURIPS2019_9015} and algorithm hyperparameters. We use the following format to describe our neural networks. The 3D CNNs are represented as $\mathrm{cnn3}\_\mathrm{l}(\mathrm{channel}\_\mathrm{size},$ $ \mathrm{kernel}\_\mathrm{size},$ $ \mathrm{stride})$, where $\mathrm{l}$ indicates the layer number. Furthermore, each 3D CNN layer is followed by non-linearity ReLU \cite{nair2010rectified}, BatchNorm \cite{ioffe2015batch}, and 2$\times$2$\times$2-MaxPool layers. The fully connected multi-layer perceptrons are described as $\mathrm{mlp}\_\mathrm{id}(\mathrm{input}\_\mathrm{size},$ $\mathrm{hidden}\_\mathrm{layer}\_\mathrm{sizes}, \mathrm{output}\_\mathrm{size})$ in which $\mathrm{id}$ indicates the network's unique identification numbers. In mlp, all hidden layers are followed by non-linearity ReLU. Finally, we use the Adam optimizer \cite{kingma2014adam} to train our neural models for $e=500$ epochs with the learning rate $5\times 10^{-4}$ and the batch size $N=64$.
\subsubsection{ScoreNet}
The network's inputs are current scene representation $S$ and viewpoint $v$. The input $S$ is processed following 3D CNN kernels $\mathrm{conv3}\_1(16, 4\times 4\times 4,2), \mathrm{conv3}\_2(32,4\times4\times4,1)$, and  $\mathrm{conv3}\_3(64,4\times4\times4, 1)$. The resulting latent representation $Z_s$ has 2880 dimensions. The viewpoint $v$ is given to $\mathrm{mlp}\_1(7, 64, 256, 512)$ to form $Z_v$. The latent space representation $Z_s$ and $Z_v$ are concatenated and passed to another $\mathrm{mlp}\_2(3392, 1024, 256, 64, 1)$ followed by a sigmoid function to obtain the predicted coverage rate $\hat{\theta} \in [0, 1]$.  
\begin{figure*}[!h]
    \centering
    \includegraphics[trim = {4cm 0cm 3cm 0cm}, clip, width = 17cm]{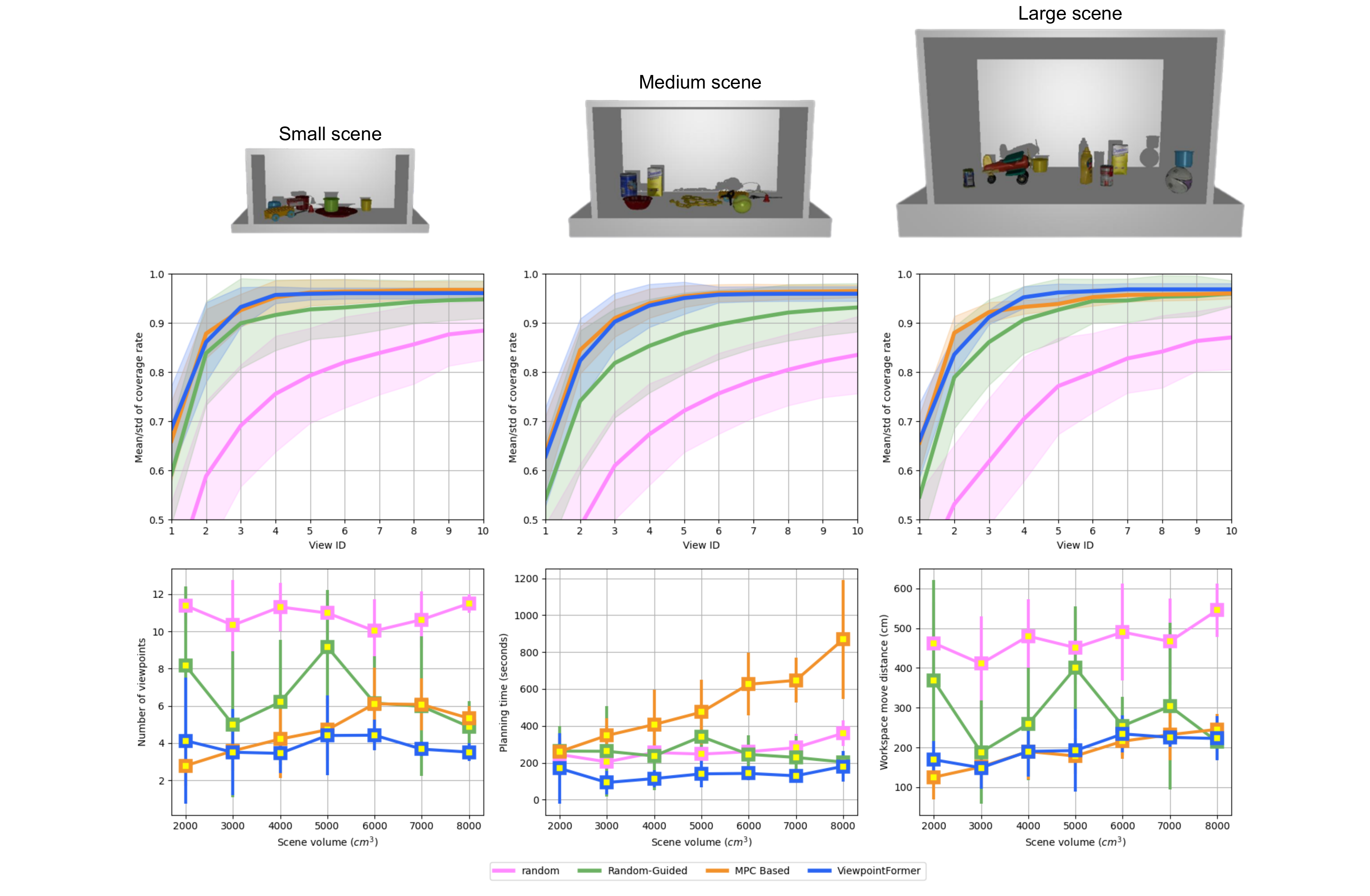}
    \caption{Statistical comparison between different active viewpoint generation algorithms. The first-row shows typical small, medium, and large sized environments. The underlying plots on the second row reflect how the coverage rate grows across different algorithms. The bottom three plots show the relationship between scene volume and the number of viewpoints, computational planning time, and the move distance in the workspace. It can be seen that our VPFormer scales very well across different environments and outperforms baseline random and random-guided methods by a significant margin.}
    \label{fig:coverage_rate_vs_view}
\end{figure*}
\subsubsection{VPFormer}
The input of VPFormer are sequences of $(c, S, v)$ tokens with a maximum length bounded below 8. We keep the architecture that transfers $S, v$ into the latent space representation $Z_S, Z_v$ the same as in the ScoreNet. $Z_S, Z_v$ and $c$ form token embedding by passing through $\mathrm{mlp}\_3(2880, 256)$, $\mathrm{mlp}\_4(512, 256)$ and $\mathrm{mlp}\_5(1, 256)$, respectively. We also apply time embedding \cite{chen2021decision} to all token sequence $(c, S, v)$ embeddings before passing them through 6 multi-headed attention-based Transformer's encoding blocks \cite{vaswani2017attention}. In each encoder block, we use eight heads each with a Dropout \cite{srivastava2014dropout} of $10\%$. A diagonal mask is applied when performing the attention calculation to ensure the current step can only access previous steps. The extracted latent representation of $v$ is then passed through $\mathrm{mlp}\_6(256, 7)$ followed by a tanh function, leading to the next best viewpoint predictions. Furthermore, during execution, we use the VPFormer to provide the first two viewpoints, and then we revert to the first stage of our Bilevel MPC algorithm for active sensing.
\subsubsection{Bilevel MPC}
We update the viewpoint distribution 5 times during the process, i.e., $N_{iter} = 5$. The initial mean $\mu$ is the viewpoint generated from the first stage, while the standard deviation $\sigma$ is set to 0.1 across all dimensions. The total number of sampled viewpoint candidates $N_{mpc}$ in each iteration is $1000$. Furthermore, we gradually decrease the number of elite samples over five iterations as $800, 500, 200, 100, 50$ to narrow our sampling distribution variance.  
\begin{table*}[!ht]
  \fontsize{7}{5}\selectfont
  \begin{center}
    \begin{tabular}{c c c c c c}
      \hline
      \multirow{2}{*}{\textbf{Viewpoint generation Algo}} & \multicolumn{5}{c}{\textbf{Metrics}}\Tstrut\Bstrut \\
       & Number of viewpoints $\downarrow$ \Tstrut\Bstrut & Success rate $(\%)$ $\uparrow$ & Planning time (s) $\downarrow$ & C-Space distance $\downarrow$ & Workspace distance $\downarrow$\\
      \hline
      Random \Tstrut\Bstrut & 11.61 $\pm$ 2.65 & 23 & 247.53 $\pm$ 78.29 & 49.34 $\pm$ 12.63 & 4.60 $\pm$ 1.08\\
      Random-Guided\Tstrut\Bstrut & 6.25 $\pm$ 3.21 & 81 & 264.82 $\pm$ 180.63 & 29.58 $\pm$ 19.33  & 2.80 $\pm$ 1.70\\
     Bilevel MPC\Tstrut\Bstrut & 5.02 $\pm$ 1.86 & \textbf{99} & 452.87 $\pm$ 212.92 & 17.70 $\pm$ 8.59 & \textbf{1.82 $\pm$ 0.64} \\
     VPFormer\Tstrut\Bstrut & \textbf{4.19 $\pm$ 1.22} & 97 & \textbf{125.72 $\pm$ 83.10} & \textbf{16.21 $\pm$ 9.34} & 1.87 $\pm$ 0.72\\
      \hline
    \end{tabular}
    \caption{This table shows the comparison between various active viewpoint generation algorithms. The VPFormer achieves the best overall performance with the minimum number of viewpoints, shortest planning time, and shortest C-space move distance. The success rate and workspace move distance of VPFormer are also near-optimal.}
    \label{tab:table1}
  \end{center}
\end{table*}
\begin{table*}[!ht]
  \fontsize{7}{5}\selectfont
  \begin{center}
    \begin{tabular}{c c c c c c}
      \hline
      \multirow{2}{*}{\textbf{Viewpoint generation Algo}} & \multicolumn{5}{c}{\textbf{Metrics}}\Tstrut\Bstrut \\
       & Number of viewpoints $\downarrow$ \Tstrut\Bstrut & Success rate $(\%)$ $\uparrow$ & Planning time (s) $\downarrow$ & C-Space distance $\downarrow$ & Workspace distance $\downarrow$\\
      \hline
     VPFormer\Tstrut\Bstrut & 4.19 $\pm$ 1.22 & 97 & 125.72 $\pm$ 83.10 & 16.21 $\pm$ 9.34 & 1.87 $\pm$ 0.72\\
     VPFormer w/o Completion\Tstrut\Bstrut & 5.03 $\pm$ 2.32 & 95 & 160.12 $\pm$ 120.67 & 18.88 $\pm$ 12.24 & 2.16 $\pm$ 1.11\\
     VPFormer w/o ScoreNet\Tstrut\Bstrut & 5.68 $\pm$ 1.95 & 97 & 162.07 $\pm$ 85.74 & 21.61 $\pm$ 9.44 & 2.29 $\pm$ 0.96\\
      \hline
    \end{tabular}
    \caption{Ablation study results show the impact of the object shape completion network and ScoreNet function on the performance of VPFormer. It can be seen that both completion and ScoreNet functions augment overall performance to some extent and lead to lower interaction steps and therefore planning times and move distances.}
    \label{tab:table2}
  \end{center}
\end{table*}
\section{Results}
\noindent In this section, we present the results and analysis of the following evaluation experiments: 1) a comparison experiment evaluating the performance of the proposed viewpoint generation algorithms against the baseline in various cluttered environments; 2) an ablation study showing the effectiveness of our various neural functions involved in the active sensing system; 3) sim-to-real transfer demonstrating our approach's performance in 10 different never-seen-before real-world cabinet settings. We use the following metrics for quantitative evaluation:\\
\begin{itemize}
    \item \textbf{Number of viewpoints:} It records how many viewpoints are proposed before achieving the desired coverage score.
    \item \textbf{Success rate:} It shows the percentage of reconstructed scenes that had maximum coverage scores above $c_{max}$ within 15 viewpoints.
    \item \textbf{Planning time:} It calculates the time consumption in computing viewpoint sequences.
    \item \textbf{Move distance:} It indicates the total distance of robot movements both in configuration space (C-space) \cite{lozano1990spatial} and work space \cite{gupta1986nature}.
    \item \textbf{Average scene coverage:} It measures how the coverage rate increases over the number of viewpoint steps.
\end{itemize}
\subsection{Comparison of Viewpoint Generation Algorithms}
\noindent We set two baseline algorithms to compare with our methods, named random and random-guided. The former method randomly samples the feasible viewpoints in the observed region while the latter algorithm selects a viewpoint from a batch of randomly sampled candidates using ScoreNet with a maximum coverage score prediction. We test all viewpoint generation algorithms in unseen 100 simulation environments. The means and the standard deviations of all metric items across different algorithms are presented in Table \ref{tab:table1}. We further divide the testing environments by volume and analyze the relationship between the performance and the scene sizes, as demonstrated in Fig \ref{fig:coverage_rate_vs_view}. The scene sizes small, medium, and large have volume $(\mathrm{cm}^3)$ in the range [2000, 4000), [4000, 6000), and [6000, 8000], respectively. The first row of figure \ref{fig:coverage_rate_vs_view} shows the typical appearance of the different-sized environments. We describe a metric-wise analysis of our results in the following. 
\subsubsection{Number of Viewpoint steps}Since our objective is to actively observe the scene using a minimal number of viewpoints, the VPFormer achieves the most satisfactory result with a mean of 4.19 viewpoints. A 4-step viewpoint sequence found by VPFormer in a narrow, cluttered environment is shown in Fig \ref{fig:sim_figure}. Rows 1-4 exhibit the robot viewpoint configurations, resulting camera observations, augmented scenes, and the corresponding completed scenes after shape inference. Each column's results come from the same viewpoint at a specific time step. We can witness that the first two viewpoints place the camera's visibility cone diagonally in the scene to cover a reasonable amount of upper space. Then, it selects viewpoints to compensate for the lower missing region. We can see that there is no unnecessary overlapping between observations, contributing to the relatively lower number of viewpoints. Furthermore, the difference between row 3 and row 4 demonstrates the effectiveness of our object completion network. The next best performance is by the Bilevel MPC method followed by the random-guided approach which requires few extra viewpoints for scene coverage. Finally, the number of viewpoints of the random approach is almost three times more than the VPFormer, reaching a mean of 11.61. The first plot in the third row of Fig \ref{fig:coverage_rate_vs_view} illustrates the number of viewpoints of all methods across different-sized environments. Overall, the results show that the VPFormer requires the least amount of viewpoints and is invariant to the scene sizes compared to other methods.
\subsubsection{Success rate} Both our VPFormer and Bilevel MPC methods exhibit around $97\%$ success rate. The former approach outputs informed viewpoints, preventing exhausting search and leading to reasonable coverage scores in few viewpoints. The latter MPC algorithm finds a locally optimal solution by iteratively optimizing the initial viewpoint distributions guided by our Scorenet, thus leading to similar success rates as VPFormer. For the random-guided algorithm, it is not likely to have viewpoint candidates directly sampled around the local optimal, resulting in a significant success rate drop, i.e., $81\%$. In a similar vein, the random method achieves only $23\%$, which is nearly four times lower than the VPFormer and the Bilevel MPC.
\begin{figure*}[ht]
    \centering
    \includegraphics[trim = {1.4cm 1.5cm 0cm 0cm}, clip, width = 18cm]{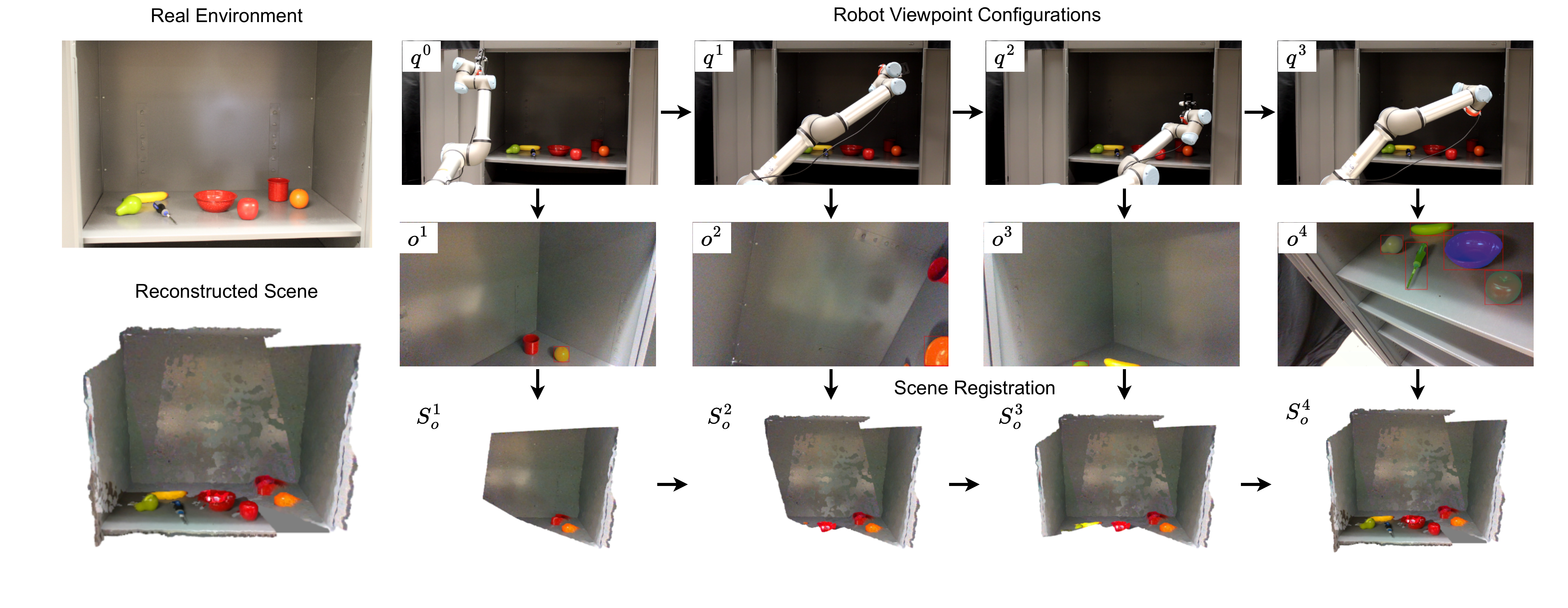}
    \caption{A viewpoint sequence generated by our VPFormer-based method to observe and reconstruct the real-world cabinet. The object setup is displayed in the top left figure. The bottom left figure is the final reconstructed scene from all viewpoints. Figures in the top row are the robot viewpoint configurations, $\{q^0, \cdots, q^3\}$, that lead to the next observations, $\{o^1, \cdots, o^4\}$. The resulting scene reconstructions, $\{S^1, \cdots, S^4\}$, after data aggregation and shape completion are shown in the bottom row. It can be seen that the reconstructed environment is realistic despite limited ambient lightening due to covered regions. }
    \label{fig:real_exp_1}
\end{figure*}

\subsubsection{Planning time} Another metric that our VPFormer achieves the best result is the planning time. For the VPFormer, only two forward passes through the neural networks are needed before outputting the viewpoint. In contrast, the computational times of our bilevel MPC algorithm are the highest among all methods since it requires an iterative optimization process to refine the initial viewpoint distribution. Finally, even though each viewpoint generation is quick for the baseline algorithms, a relatively low success rate and a high number of viewpoints contribute to its prolonged planning time. Regarding scalability over scene size, the second plot on the third row of Fig. \ref{fig:coverage_rate_vs_view} compares all methods. It can be seen that the VPFormer outperforms all other methods in planning time. 
\subsubsection{Move distance} The last plot of Fig. \ref{fig:coverage_rate_vs_view} presents move distance of all methods. The VPFormer and bilevel MPC algorithms share similar move distances in the C$-$space and workspace. A smaller move distance allows the robot manipulator to get to the destination much faster and, as a result, boosts the performance of the active sensing process. Furthermore, the baseline algorithms, random-guided and random, select many unnecessary viewpoints, leading to an average move distance at least two times larger than our proposed methods both in C-space and workspace. Note that the move distance strongly correlates with the number of viewpoints. Since the VPFormer has relatively stable viewpoint numbers, its planning time is also bounded in a narrow range across all different and variable size environments.

\subsubsection{Average coverage rate} The relationship between average coverage rate versus view ID is shown in the second row of plots in Fig \ref{fig:coverage_rate_vs_view}. Our VPFormer and bilevel MPC  algorithm can get an average coverage rate above $90\%$ in three views despite the scene volume. It takes the random-guided method, on average, two more viewpoints to get to the same coverage rate, while the random algorithm rarely reaches $90\%$ in all environments.\par
In conclusion, the comparison analysis of all method in unseen simulation environments highlights that our VPFormer achieves the best performance with the lowest number of views, planning time, and C$-$space move distance. Its results in the success rate and the workspace move distance are also very close to the optimal.
\begin{figure*}[ht]
    \centering
    \includegraphics[trim = {1cm 1.5cm 0cm 11cm}, clip, width = 18cm]{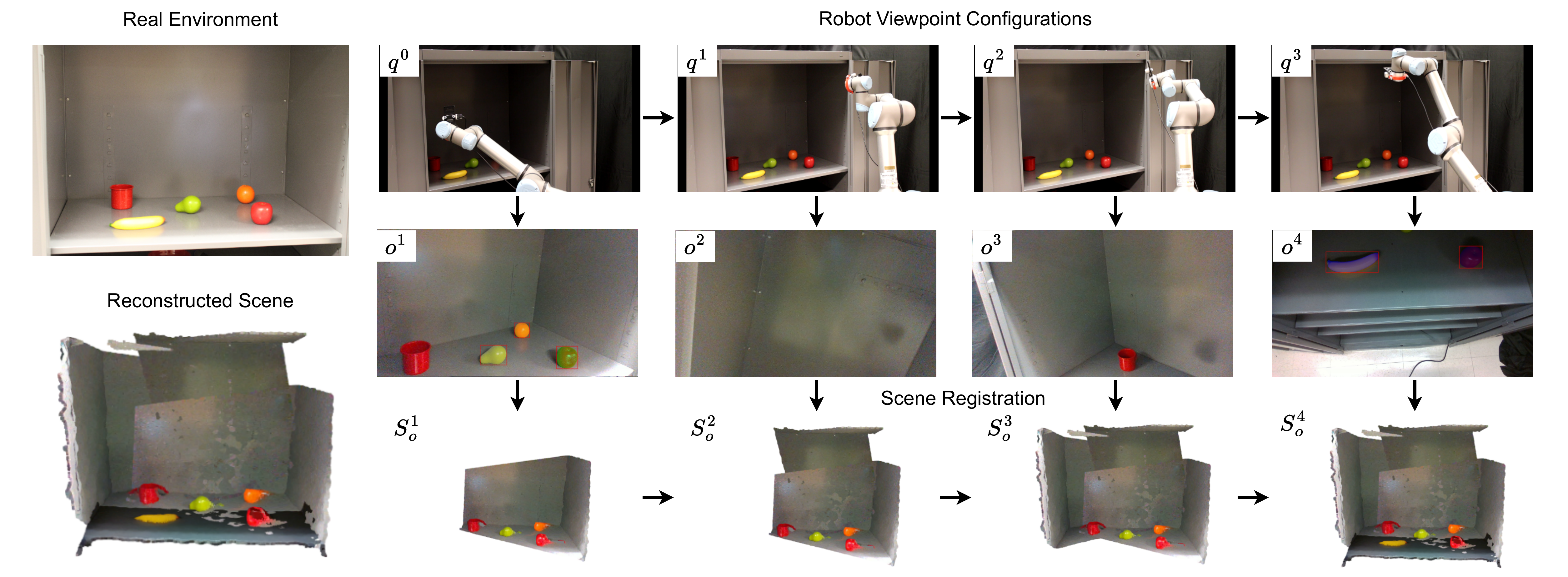}
    \caption{This figure demonstrates another viewpoint sequence by VPFormer-based active system. Note that the scene distortion is due to real sensor noise, which causes erosion around object edges. }
    \label{fig:real_exp_2}
\end{figure*}
\subsection{Ablation Studies}
\noindent We conduct ablation studies to show the importance and effectiveness of our shape completion network and the ScoreNet. The results are summarized in the Table \ref{tab:table2}. 
\subsubsection{Shape Completion Network}In this experiment, we evaluate the influence of the shape completion network on our VPFormer. We perform evaluations on our simulated test dataset using the VPFormer without the shape completion network in the scene registration pipeline. Therefore, the extracted object point clouds from new observation are registered directly into the scene without passing them through our shape completion module. The experiment result reflects that VPFormer takes an additional viewpoint when executed without the object completion network. As a result, the planning time and moving distance both in C-space and workspace also increase. However, the success rate does not change as the original VPFormer has a significant performance margin towards the 15 viewpoints limit. Hence, although the completion network marginally enhances VPFormer performance, it plays a positive role in overall scene reconstruction efficiency. 
\subsubsection{ScoreNet and VPFormer} In this experiment, we directly use the VPFormer output instead of generating more samples around the predicted viewpoint for further refinement via ScoreNet predictions. The results indicate that the viewpoint number increases by 1.5 on average, but the success rate remains stable. Furthermore, the planning time and move distance also increase slightly. Therefore, this experiment shows the importance of ScoreNet in further refining the neural-generated viewpoints. However, we should also note that since VPFomer performance only drops slightly, the results also demonstrate its ability to perform near-optimal viewpoint planning in cluttered environments. 

\subsection{Real Experiment}
\noindent In this section, we evaluate the sim-to-real transfer capabilities of our VPFormer framework. We create ten scenes with different object configurations. Our method exhibited a 100\% success rate in scene coverage, with an average number of viewpoint steps being 5.3. The two viewpoints sequences found by the system using the VPFormer are shown in Fig. \ref{fig:real_exp_1} and \ref{fig:real_exp_2}.  Both figures show the actual scene configuration and the final scene reconstruction on the left side, followed by a detailed display of the viewpoint sequence. The rows reveal the robot viewpoint configurations, the camera observations, and the reconstructed scene at different time steps. Furthermore, in these two experiments, the scene coverage takes up to 4 viewpoints. Although the real experiments are successful, we still find two major challenges. First, depth measures from the camera can be inaccurate along the edge of objects, resulting in distorted objects in the scene representation. Therefore, we applied the image erosion \cite{sreedhar2012enhancement} on the segmented image to reduce distortions before performing the object completion process. Another major challenge of the real experiments is the performance of instance segmentation algorithms. Even with the most advanced method like Detectron2 \cite{wu2019detectron2}, there is still no guarantee that all objects captured in the image will be correctly extracted and segmented. Our completion network does not complete the undetected objects, which results in slightly more viewpoints.   

\section{Discussion}
\noindent We presented a novel robot manipulator-based active neural sensing approach that works in unknown, cluttered environments. We show our system can explore the unknown setups using a minimal number of viewpoints in both simulated and real-world narrow, cabinet-like scenarios. The results exhibit high performance compared to traditional baselines in terms of the number of viewpoints, scene coverage success rates, and overall planning time. Although, unlike prior work, our approach generalizes to complex cluttered, real-world scenarios, it suffers from inaccurate depth camera measurements and the unsatisfactory performance of existing instance segmentation algorithms. Therefore, in our further studies, we plan to improve the quality of our scene representation by introducing novel object matching methods robust to noisy depth measurements of real camera sensors. Furthermore, in the future, we also aim to investigate the applications of our proposed framework in search and rescue tasks using snake-like robots as we believe those scenarios can directly employ our current method for disaster sites' active sensing.

\bibliographystyle{IEEEtran}
\bibliography{./citation.bib}{}


 





\end{document}